# Iterative self-transfer learning:
# A general methodology for response time-history prediction based on small dataset

Yongjia Xu,[a] Xinzheng Lu,[b,1] Yifan Fei,[c] Yuli Huang[b]

[a] *Zhejiang University–University of Illinois at Urbana Champaign Institute, Zhejiang University, Zhejiang, China, 314400*

[b] *Department of Civil Engineering, Tsinghua University, Beijing, China, 100084*

[c] *Beijing Engineering Research Center of Steel and Concrete Composite Structures, Tsinghua University, Beijing, China, 100084*

## Abstract

There are numerous advantages of deep neural network surrogate modeling for response time-history prediction. However, due to the high cost of refined numerical simulations and actual experiments, the lack of data has become an unavoidable bottleneck in practical applications. An iterative self-transfer learning method for training neural networks based on small datasets is proposed in this study. A new mapping-based transfer learning network, named as deep adaptation network with three branches for regression (DAN-TR), is proposed. A general iterative network training strategy is developed by coupling DAN-TR and the pseudo-label (PL) strategy, and the establishment of corresponding datasets is also discussed. Finally, a complex component is selected as a case study. The results show that the proposed method can improve the model performance by near an order of magnitude on small datasets without the need of external labeled samples, well behaved pre-trained models, additional artificial labeling, and complex physical/mathematical analysis.

---

[1] Corresponding author at luxz@tsinghua.edu.cn

# 1. Introduction

The response time-history prediction of structures is an essential task in civil engineering. Simulation based on refined models has advanced significantly in recent years. However, the complexity of the model has risen dramatically at the same time, resulting in a substantial increase in computational resources and time spent on simulation tasks.

In recent years, data-driven surrogate modeling for response time-history prediction tasks (referred to as "the target task" in the following sections), particularly those based on deep neural networks, has received much attention (e.g., Sabokpa et al., 2012; Zopf & Kaliske, 2017; Wang et al., 2020; Xu et al., 2020, 2021; Lu et al., 2021a). The deep neural network-based surrogate models have several advantages, including a strong nonlinear fitting ability, end-to-end (without human management during intermediate processes), and low burden in terms of physical/mathematical analysis. However, network training is always performed based on large datasets. Completing many analyses based on refined physics-based models or experiments is extremely costly. Lacking of data has become a major impediment.

Transfer learning is an important strategy to compensate for the lack of data. Mining and transferring knowledge from the source domain can aid in the completion of a task in the target domain. Transfer learning has been widely recognized in many fields such as computer vision and natural language processing (e.g., Gopalakrishnan et al., 2017; Devlin et al., 2019; Brown et al., 2020). Many researchers have turned their attention to transfer learning in the field of generalized response prediction, particularly those related to damage detection and monitoring (e.g., Gao & Mosalam, 2018; Gopalakrishnan et al., 2017). Other related studies have focused on crack propagation simulation (e.g., Goswami et al., 2020), signal recognition (e.g., Titos et al., 2019; Lu et al., 2021b), building identification or instance segmentation (e.g., Kim et al., 2018; Shen et al., 2021; Suh et al., 2022), and model reconstruction (Lee et al., 2022).

However, it is currently difficult to construct a general source domain for transfer learning in the response time-history prediction field. In particular, the input and output have specific physical meanings, and each type of structure has unique response characteristics, resulting in a distinct and fixed mapping relationship between the input and output (rather than the probability mapping in language models). Consequently, widely used large-scale datasets for transfer learning, such as those adopted by BERT and GPT in natural language processing (Devlin et al., 2019; Brown et al., 2020), ImageNet and CIFAR-10 in computer vision (Deng et al., 2009; Krizhevsky, 2009), are not ideal source domains in this field. Furthermore, to the authors' best knowledge, no reliable unsupervised source domain establishment technology exists at present.

For all types of transfer learning methods, the aforementioned source domain establishment problem is unavoidable. The pseudo-label (PL) strategy is a semi-supervised learning strategy with high universality and reliability (Lee, 2013; Loog, 2015). However, when using the PL strategy, the initial model should label most of the samples accurately. Otherwise, serious model drift issues may arise. Meanwhile, the performance improvement resulted by iteratively using the PL strategy is expected to decay continuously based on the low-density separation assumption and entropy regularization theory (Lee, 2013). Therefore, it is necessary to develop technologies that can "generalize" the features extracted by networks.

The transfer learning technology can extract generalized features between the source and target domains. Several researchers are currently using this strategy in conjunction with the PL (e.g., Zhang et al., 2015; Yan et al., 2017; Xia et al., 2021; Isobe et al., 2021). However, to obtain the pseudo-labels of the target domain samples, these studies still rely on source domain datasets with sufficient samples (or well behaved pre-trained models), and generally make use of the inherent information from the category labels. As mentioned previously, there is a lack of universal source domain and the sequences do not have category labels (a regression problem) in the target task, and it is impossible to distinguish the samples from the input only (see Section 2.2). Consequently, the methods mentioned above are inapplicable.

Considering the natural similarity between the pseudo-labeled and the real samples, it is conducive to the realization of positive transfer. Meanwhile, compared with the labeled dataset, a pseudo-labeled dataset with inherent systematic deviation and random noise has different eigenspace. As a result, iteratively using the PL strategy to construct the source domain is feasible. Therefore, this study proposes a general transfer-learning framework that is suitable for response time-history prediction based on small datasets. This framework solves the problem of establishing the source domain in a novel manner by iteratively applying PL strategies and generalizing the network characteristics using transfer learning technology, which can significantly improve the model performance based on small datasets only.

## 2. Key technologies

## 2.1 Supporting Technology

### 2.1.1 Deep adaption network (DAN)

Tan et al. (2018) classify the deep transfer learning techniques into four main categories: instance-based (e.g., Yao & Doretto, 2010; Xu et al., 2017), mapping-based (e.g., Tzeng et al., 2014, 2015; Long et al., 2015, 2017a), network-based (e.g., Yosinski et al., 2014; Brown et al., 2020; Vania & Lee, 2021), and adversarial-based (e.g., Ganin et al., 2016; Yu et al., 2019). For the target task, the labeled samples are high-dimensional and limited by the aforementioned data acquisition problem. These characteristics may make the determination of sample similarity and selection of appropriate weights difficult. Furthermore, owing to overfitting, models with reasonable generalization abilities are difficult to be obtained in fine-tuning process and adversarial discriminant network training. As a result, mapping-based transfer learning technology is chosen for this research.

A typical mapping-based transfer learning technique is the DAN (Long et al., 2015). The features extracted by the first several layers of the network are more transferable, whereas those extracted by the upper layers are more task-specific (Yosinski et al., 2014). As a result, the parameters of the network's first several layers in DAN are directly shared between the source and target domains, while the parameters of the upper layers are constrained by minimizing the multi-kernel maximum mean discrepancy (MK-MMD). The MMD method uses the kernel-mapping algorithm to map the features of both the source and target domains into the high-dimensional reproducing kernel Hilbert space and computes the distances between the embedding results. MK-MMD is an extension of MMD that employs multiple kernels to better approximate feature space distributions (Cheng et al., 2020). This architecture is ideal for extracting generalizable features and achieving positive transfer. Because the basic principles of many architectures proposed in other studies (e.g., the deep domain confusion (DDC) network, joint CNN, and joint adaptation network (JAN) (Tzeng et al., 2014, 2015; Long et al., 2017a)) are similar to that of DAN, only DAN is discussed in detail herein.

The loss function of DAN comprises two parts (Equation (1)): cross-entropy classification loss and MK-MMD loss. The first part is calculated using labeled samples from the source ($\mathbf{X}_s$ and $\mathbf{Y}_s$) and target domains ($\mathbf{X}_{t,L}$ and $\mathbf{Y}_{t,L}$, assuming that $\mathbf{Y}_{t,L}$ is available). Long et al. (2015) provide a more detailed introduction to the network architecture and loss calculation.

$$Loss = Loss_{\text{cls}} + \lambda \cdot Loss_{\text{MK-MMD}} \qquad (1)$$

where, $Loss_{\text{cls}}$ represents the cross-entropy loss of the classification results; $Loss_{\text{MK-MMD}}$ represents the MK-MMD among the hidden representations from the tailored nets of the source and target domain; $\lambda$ is the weight coefficient.

### 2.1.2 PL strategy

Several data augmentation strategies, such as RandAugment, Mixup, and MixMatch, have been proposed in previous studies (e.g., Zhang et al., 2017; Berthelot et al., 2019; Cubuk et al., 2020). However, the target task is a high-dimensional time-series regression task with strong nonlinearities, which limits the use of many data augmentation strategies. Taking concrete components as an example, because of the historical dependency and inconsistent tension/compression behavior of it, sequence scaling/cropping and combining of different samples (including the RandAugment and Mixup) are unreliable.

The PL strategy is a highly adaptable semi-supervised data augmentation strategy that works with a wide range of input and output data. Existing studies have demonstrated the logic of this technique in terms of clustering assumptions and entropy regularization (Lee, 2013). Key steps of adopting PL strategy include:

(1) Train the network models with labeled samples

(2) Provide pseudo-labels to unlabeled samples

(3) Combine the pseudo-labeled and labeled samples to train new models.

When the sample space (eigenspace) is uniformly and densely filled, the PL strategy could not yield further performance improvement. The case study in Section 4 also shows that the PL technique can be employed iteratively (one to two times in general), while for pursuing continuous improvement, it is necessary to integrate the PL strategy with transfer learning. In addition, the mean-teacher technique is used during the training process, which is a widely acknowledged data augmentation strategy based on consistency (Tarvainen & Valpola, 2017).

## 2.2 DAN with three branches for regression (DAN-TR)

### 2.2.1 Analysis of existing networks

After examining the characteristics of the target task, it is discovered that the DAN and similar architectures could not be used directly for the following reasons.

(1) The networks mentioned above are intended for domain adaption tasks. The conditional distributions of the source and target domains are assumed to be the same ($Q_s(\mathbf{Y}_s|\mathbf{X}_s) = Q_t(\mathbf{Y}_t|\mathbf{X}_t)$) (Long, 2014). However, in the target task, the marginal distributions of the source and target domains are the same ($P_s(\mathbf{X}_s) = P_t(\mathbf{X}_t)$), but the conditional distributions are inconsistent ($Q_s(\mathbf{Y}_s|\mathbf{X}_s) \neq Q_t(\mathbf{Y}_t|\mathbf{X}_t)$), which shows the characteristics of "multi-task learning" problems to some extent (Long, 2014). Meanwhile, there is only one task in essence, as the source domain is constructed entirely based on the target domain.

Taking the seismic analysis of the components as an example, the ground motion input of two different components can be any ground motion ($P_s(\mathbf{X}_s) = P_t(\mathbf{X}_t)$). However, even if the same ground motion ($\mathbf{X}_s = \mathbf{X}_t$) is taken as the input, the responses of the two components obviously differ from each other ($Q_s(\mathbf{Y}_s|\mathbf{X}_s) \neq Q_t(\mathbf{Y}_t|\mathbf{X}_t)$).

(2) The shared network is composed of CNN and fully connected layers because the aforementioned networks are designed for computer vision tasks. Some networks also consider the degree of similarity between the category labels. However, because the target task is a high-dimensional time-series regression task, CNN is not an ideal basic network, and no category labels exist.

(3) The aforementioned networks are designed for tasks with sufficiently labeled source domain samples (or well behaved pre-trained models). Consequently, it has a low reliance on labeled samples in the target domain (the source domain can provide valuable information). As previously stated, the source domain in the target task is built based on very limited target samples and the PL strategy. Compared with real labels, it consists of less effective information. Besides, unlabeled samples do not contain any domain information. In other words, the source and target domains cannot be distinguished from the input standpoint. Therefore, labeled samples in the target domain have higher importance and must be comprehensively used.

### 2.2.2 Design of the DAN-TR network

Based on the above analysis, the DAN-TR network is proposed (Fig. 1), which has good adaptability to the target task:

(1) Two branches with parameter sharing at the first several layers are established for pseudo-labeled samples in the source domain ($\mathbf{X}_s \rightarrow \mathbf{Y}_s$) and labeled samples in target domain ($\mathbf{X}_t \rightarrow \mathbf{Y}_t$), respectively. These two branches are designed referring to the principles of multi-task learning network architectures (Long et al., 2017b; Ma et al., 2018).

(2) Referring to the architecture of the DAN (Long et al., 2015), a domain adaptation branch ($\mathbf{X}_t \rightarrow \mathbf{Y}_{st}$) is added, which shares the parameters with the source domain branch ($\mathbf{X}_s \rightarrow \mathbf{Y}_s$). Meanwhile, because the source and target domains have the same marginal distribution ($P_s(\mathbf{X}_s) = P_t(\mathbf{X}_t)$), the unlabeled samples do not contain any domain-specific information. Consequently, the adaptive branch uses the labeled samples in the target domain ($\mathbf{X}_t$) as the input instead of the unlabeled samples ($\mathbf{X}_{un}$) used in the DAN.

(3) The MK-MMD loss (Long et al., 2015) between the tailored nets of the adaptation branch ($\mathbf{X}_t \rightarrow \mathbf{Y}_{st}$) and target domain branch ($\mathbf{X}_t \rightarrow \mathbf{Y}_t$) is calculated. The network parameters are optimized through the backpropagation of the MK-MMD loss to improve the transferability and generalization ability of the extracted features. The output of the domain adaptation branch $\widehat{\mathbf{Y}}_{st}$ does not participate in backpropagation to avoid the convergent of branch optimization directions, which may have a negative impact on domain adaptation.

(4) The shared net is made up of long short-term memory neural network (LSTM) layers (Hochreiter & Schmidhuber, 1997), and the tailored net comprises a combination of LSTM and fully connected layers. The LSTM network is well suited for extracting and mapping features from time-series data, and it has been widely used in many similar tasks (e.g., Graves, 2012; Chen et al., 2020a; Xu et al., 2021).

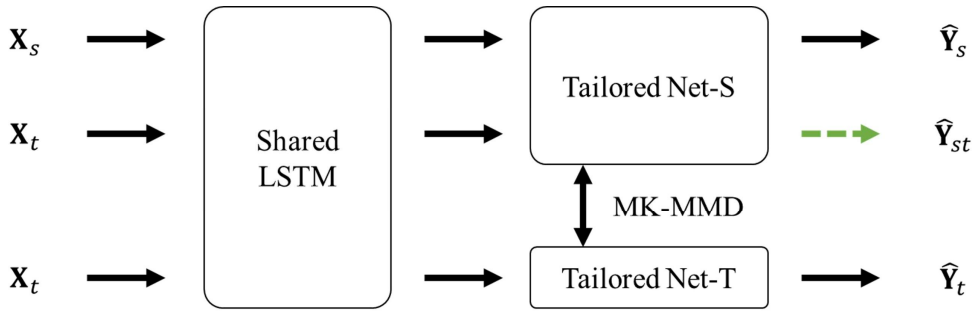

Fig. 1: Architecture of the proposed DAN-TR

The loss of the DAN-TR network includes the regression loss and MK-MMD loss (Equation (2)):

$$Loss = Loss_{\text{reg}} + \lambda \cdot Loss_{\text{MK-MMD}} \tag{2}$$

where, $Loss_{\text{reg}}$ represents the regression loss, which can be calculated using Equation (3); $Loss_{\text{MK-MMD}}$ and $\lambda$ can be calculated using Equations (4) and (5) (Long et al., 2015; Wang, 2021), respectively.

$$Loss_{\text{reg}} = \frac{1}{n_{\alpha s}} \sum_{i=1}^{n_{\alpha s}} J_R(\theta_s(x_i^{\alpha s}), y_i^{\alpha s}) + \frac{1}{n_{\alpha t}} \sum_{i=1}^{n_{\alpha t}} J_R(\theta_t(x_i^{\alpha t}), y_i^{\alpha t}) \tag{3}$$

$$Loss_{\text{MK-MMD}} = \sum_{l=l_1}^{l_2} f_{\text{MK-MMD}}(H_{ls}, H_{lt}) \tag{4}$$

$$\lambda = \frac{2}{1 + e^{\frac{-10 \cdot n_b}{N}}} - 1 \tag{5}$$

where, $n_{\alpha s}$ and $n_{\alpha t}$ denote the number of labeled (pseudo-labeled) samples in the source and target domains, respectively; $\theta_s$ and $\theta_t$ denote the network parameters of the source and target domain branches; $J_R$ denotes the mean square error (MSE) loss function; $x_i$ and $y_i$ denote the input and output of the $i$-th sample (in the source domain or the target domain); $f_{\text{MK-MM}}$ is the MK-MMD loss function (Long et al., 2015); $H_{ls}$ and $H_{lt}$ represent the hidden layer states of the source domain and target domain branches, respectively; $l_1$ and $l_2$ represent the start and end of the layer series participating the MK-MMD calculation; $n_b$ is the current number of conducted training step; $N$ is the total number of training steps in current task.

## 3. Framework

### 3.1 Establishment of labeled and unlabeled datasets

3.1.1 Labeled dataset (target domain)

The target domain (denoted as $\mathbf{D}_t$) is a small dataset with real labels that must be established based on refined physical models or real experiments. Obtaining a large number of labeled samples in practical applications is difficult owing to the high experimental or computing costs. Thus, the number of target domain samples is always limited. The validation and testing sets are always selected from labeled samples to ensure the effectiveness of the model performance evaluation.

3.1.2 Unlabeled dataset (source domain)

The large-scale unlabeled dataset (denoted as $\mathbf{D}_u$) is made up of samples with only inputs (without response, i.e., not labeled). For example, when using ground motion accelerations as the input to predict structural displacements, samples can be directly downloaded from open-access ground motion databases such as the PEER NGA (PEER, 2021) and K-NET (NIED, 2021). If displacement is regarded as the input to predict the reaction force, displacement sequences can be obtained by integrating the ground motions. Parts of the unlabeled dataset can be added to the training set of subsequent iterations after obtaining pseudo-labels (denoted as $\mathbf{D}_f$). There is no need to divide the unlabeled dataset further.

It is important to note that the primary goal of this study is to reduce the cost of creating datasets. Furthermore, the marginal distributions of the source and target domains are the same ($P_s(\mathbf{X}_s) = P_t(\mathbf{X}_t)$), so the unlabeled samples do not need to contain any domain information. Therefore, high-efficiency and low-cost methods should be adopted to create an unlabeled dataset. It is not necessary to perform a detailed physical/mathematical analysis of the target task or to focus on improving the representativeness of the samples. Only ostensible consistency is required. Many existing studies have recognized and applied similar concept of using unlabeled data or generating samples based on random noise (e.g., Ganin et al., 2016; Devlin et al., 2019; Berthelot et al., 2019; Cubuk et al., 2020).

## 3.2 Training framework

The model-training framework comprises the following five key steps (Fig. 2):

**(1) Step 1:** Train the initial model (Model-I) on the target domain $\mathbf{D}_t$

The performance of Model-I is likely to be unsatisfactory under small sample conditions. Its main role is to be used as an initial model for constructing pseudo-labeled datasets.

**(2) Step 2:** Build a pseudo-labeled dataset (denoted as $\mathbf{D}_{f,i}$, $i$ is the number of iterations) based on Model-I using the PL strategy, and train neural networks on the dataset $\mathbf{D}_{f,i}$ for several iterations to obtain Model-F$_i$ (one or two iterations are generally sufficient).

There are significant deviations between the pseudo and real labels. However, compared to models trained directly on $\mathbf{D}_t$, advantages exist in terms of accuracy and stability.

① The PL strategy can improve the model accuracy by increasing the sampling density in the sample space (eigenspace) and adding perturbations to enhance feature universality.

② When the number of training samples is limited, the network optimization direction is highly erratic, and the performance of different models can differ significantly. The stability of the training will be significantly improved after enlarging the training set using the PL strategy, thus reducing the need for repeated training.

**(3) Step 3:** Transfer learning is performed between the target domain $\mathbf{D}_t$ and the most recent pseudo-labeled source domain $\mathbf{D}_s$ (obtained by Model-F$_i$) using the DAN-TR. The obtained model is denoted as Model-TR$_i$.

At first, all samples in the target domain are labeled. After several iterations, the target domain can be enlarged with relatively high-quality pseudo-labeled samples. Based on the preceding analysis:

① Transferable knowledge between the source and target domains can be extracted using domain adaptation technology, resulting in improved model performance in the current iteration.

② Feature generalization increases the size of the eigenspace generated by the PL strategy in step 2, making the samples appear "sparse" in a larger eigenspace. As a result, this step increases the potential for improving the model accuracy in subsequent iterations.

**(4) Step 4:** Iteratively utilize steps 2 and 3 until the model performance cannot be improved any further (or the anticipated performance is achieved)

**(5) Step 5:** Train models on the final dataset $\mathbf{D}_{last}$ and choose the best model (Model-L) for practical applications.

It is clear that the proposed self-iterative transfer learning method differs significantly from existing methods such as traditional PL strategy, consistency-based methods, pre-trained and fine-tune strategies, and conventional data augmentation. It innovatively integrates the PL strategy with the proposed DAN-TR network and establishes an iterative training framework. Because of the strong nonlinearity discussed in Section 2.1, traditional data augmentation methods are not used in this framework. Meanwhile, when compared to the pre-train and fine-tune strategy, the proposed method eliminates the need for large-scale self-labeled source domains, which can be difficult to be built efficiently in time-history prediction tasks. In addition, the consistency loss is adopted to enhance the stability of the obtained models, yet it's not the core of the proposed method. In Section 4.5, a comparison between the proposed method and existing methods is provided.

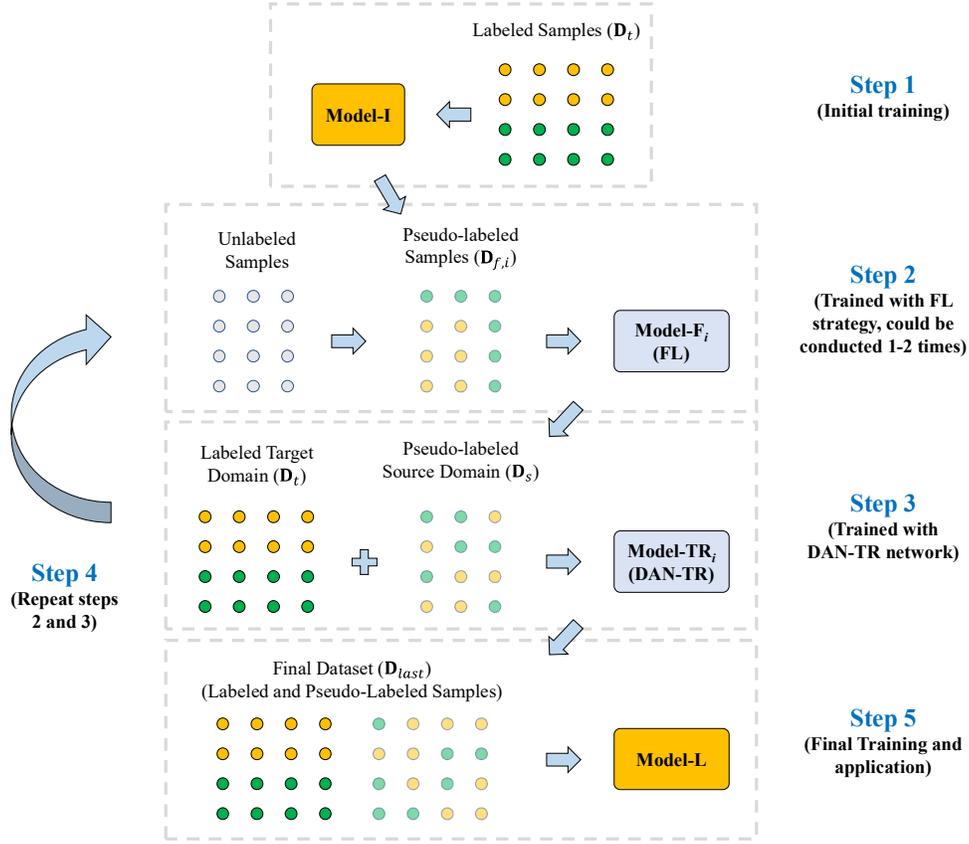

Fig. 2: Framework of the training process

It should be noted that, in steps 1 to 4 of the training process, there is no need to adopt a very complex network. The self-transfer process can be efficiently implemented with simplified networks (such as the LSTM). For practical applications, the network hyperparameters can be adjusted further in step 5. A more complex network (such as GRU + Attention (GA) network proposed by Wang et al., 2020) can be adopted in the final training (step 5) based on the constructed dataset (including labeled and pseudo-labeled samples). Detailed discussions will be carried out in Section 4.6.

## 4. Case study

### 4.1 Brief introduction

4.1.1 Introduction to the refined finite element (FE) model

A refined FE brace model established in LS-DYNA (Uriz & Mahin, 2008; Huang, 2009) is selected as the case study (Fig. 3). A calibrated cyclic damage plastic model and multiscale meshed fully-integrated shell elements are adopted to simulate the complex behavior of the brace component under hysteretic loads, which includes material nonlinearity, geometric nonlinearity, degradation, damage, and fracture evolution. This model has been well validated and widely applied in the analysis of various structures and components (e.g., Li et al., 2013; Lai et al., 2015; Moharrami & Koutromanos, 2017), and is considered as one of the representative refined models with high complexity.

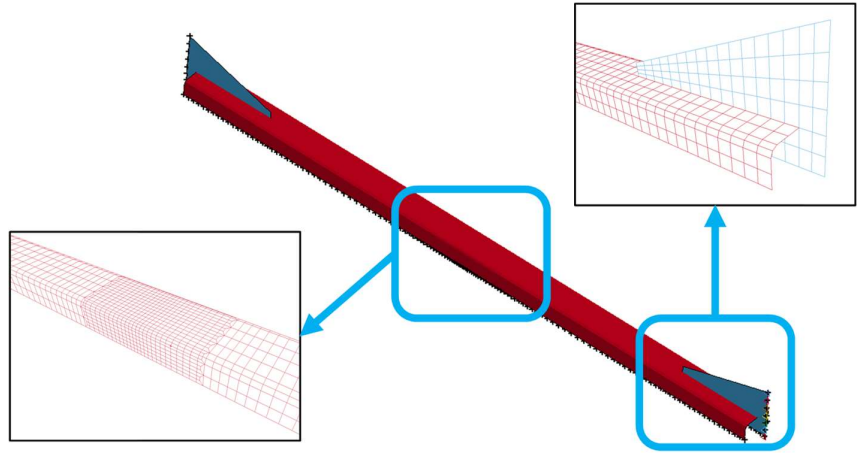

Fig. 3: Refined FE model of brace in LS-DYNA (half-structure based on symmetric boundary condition; fixed boundary on the left side and uniaxial loading on the right side)

### 4.1.2 Establishment of labeled dataset

Following the creation of the refined FE model, 200 ground motion records with various characteristics are selected from the K-NET database (NIED, 2021), and displacement sequences of them are obtained using the average acceleration integration. In addition, 200 artificial displacement sequences are generated based on the wave superposition of sine waves of different periods to increase dataset diversity. The amplitudes of these sequences are fed into the refined FE model as inputs. 400 pairs of displacement-reaction force time-history are obtained after the simulation based on the explicit integration method.

Fig. 4 shows the peak value distributions of the input (Fig. 4a) and output (Fig. 4b) sequences in the labeled dataset. The amplitudes of input sequences are relatively evenly distributed between 3.0-3.5 inch, and that of the output sequences ranges from 243-266 klbf. The simulation process can cover various stages of the component (from the elastic stage to fracture).

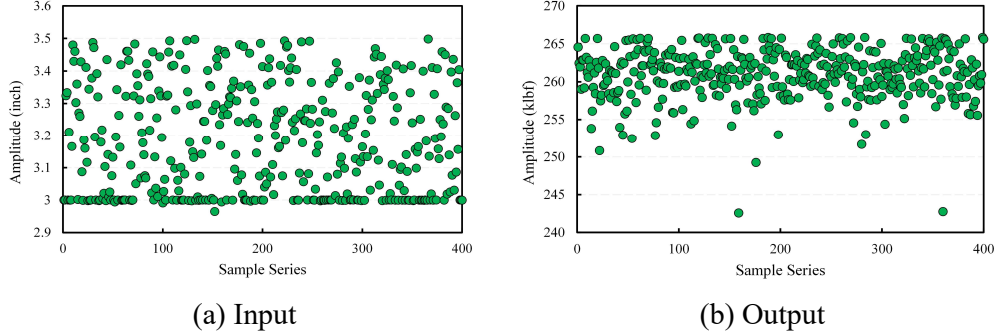

<div align="center">(a) Input            (b) Output</div>

Fig. 4: Peak value distributions of input and output sequences

These 400 pairs of samples are min-max normalized to the range of [-1, 1] (Zhang et al., 2019). Subsequently, all the data are randomly divided into training, validation, and testing datasets, with 320 (80%), 40 (10%), and 40 (10%) samples, respectively. The validation and testing sets will remain unchanged throughout the case study, whereas the small-scale training datasets required in the following sections will be sampled from the 320 samples in the original training dataset (named as $\mathbf{D}_{320}$).

The average calculation time in LS-DYNA based on the refined FE model ranges from 25.80-29.13 h. (platforms: Intel Xeon CPU E5-2695 v4 @ 2.10 GHz or Intel Xeon CPU E5-2650 v3 @ 2.30 GHz). Consequently, the dataset establishment process is expensive, thus the development of response time-history prediction methods suitable for small-scale datasets is very meaningful.

### 4.1.3 Unlabeled dataset construction

Theoretically, an unlabeled sample can be any displacement sequence. Consequently, varied ground motions are chosen from the open-access databases (PEER, 2021; NIED, 2021), and corresponding displacement sequences are obtained through integration. To generate more samples, the obtained displacement sequences are sliced, spliced, or averaged with random weights. Finally, an unlabeled dataset with 140,000 input sequences is created. This is an efficient process that requires only a few minutes. The unlabeled data do not need to be generated based on physical/mathematical analysis, nor do they need to be highly representative, as stated in Section 3.1. Only ostensible consistency is necessary, that is, all inputs are displacement sequences.

## 4.2 Hyperparameter selection

The settings of the network architectures and important hyperparameters are discussed in this section, as listed in Table 1. In real applications, generating a dataset with hundreds of samples through refined simulations is not cost-effective. Therefore, the dataset $\mathbf{D}_{320}$ is only used as a comparison group in this study. During the training process, the model parameters are independently initialized three times to reduce the influence of randomness. The final model performance is determined using the average MSE loss of the three models.

Table 1: Settings of network architectures and important hyperparameters

| Network | Definition | Value | Reference / Discussion |
|---|---|---|---|
| DAN-TR (Shared LSTM) | The number of shared LSTM layers | 2 | Zhang et al. (2019), and compare the values of 1-3 |
| | The dimension of hidden state (the same in LSTM layers and fully connected layers) | 128 | Compare the values of 64, 100, 128, 150, and 200 in terms of accuracy and training efficiency |
| DAN-TR (Tailored Net (S/T)) | The number of LSTM layers in tailored nets | 2 | Refer to Zhang et al. (2019) and compare the values of 1, 2 and 3 |
| | The number of fully connected layers (including the last output layer) | 2 | Refer to Long et al. (2015) and Yosinski et al. (2014) |
| LSTM (training based on PL) | The number of LSTM layers | 3 | Refer to Zhang et al. (2019) and compare the values of 2, 3 and 4 |
| | The number of fully connected layers (including the last output layer) | 2 | Refer to Zhang et al. (2019) |
| | The dimension of hidden state (the same in LSTM layers and fully connected layers) | 200 | Determined based on Xu et al. (2022) |
| | The weight adopted when updating the parameters of the teacher model | 0.999 | Tarvainen & Valpola (2017) |
| Common Settings | The value of learning rate | - | Dynamically adjusted in each training step |
| | The type of activation function adopted in all networks | ReLU | ReLU activation function is widely recognized |
| | The optimization algorithm | Adam | Kingma & Ba (2014) |
| | The number of selected samples in the pseudo-labeled dataset in each iteration | 15,000 or 30,000 | Compare the values of 10,000, 15,000, 30,000, 60,000, and 140,000 |

### 4.3 Training and validation on small-scale datasets

To verify the benefits of the proposed method on small-scale datasets, training dataset $\mathbf{D}_{10}$ is created by randomly selecting 10 samples from all 320 training samples, and the proposed training process is carried out. Table 2 lists the validation MSE in each iteration. Fig. 5 depicts the relative reduction of the average MSE loss in each iteration (compared to the previous iteration). Iterations $A$-1 and $A$-2 represent the first and second attempts in iteration $A$, respectively. The tentative training (3-1, 9-1, and 11-1 in Table 2) is used to confirm that if only the PL strategy is used iteratively, model performance improvement will exponentially decay. Iteration $A$+1 is carried out based on attempt $A$-2 (transfer training based on the DAN-TR network). The models and results of training attempts $A$-1 are only used for discussion and will have no impact on subsequent iterations. These training attempts are not necessary to be included in practical applications.

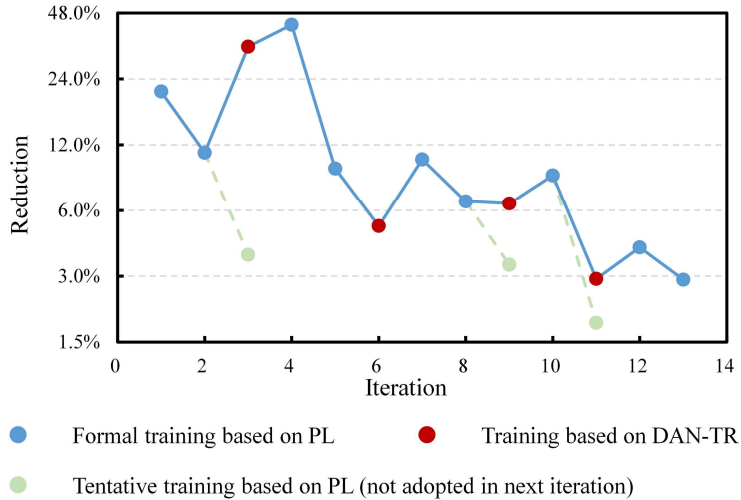

- ● Formal training based on PL
- ● Training based on DAN-TR
- ● Tentative training based on PL (not adopted in next iteration)

Fig. 5: Relative reduction of MSE Loss in each iteration ($\mathbf{D}_{10}$)

Table 2: Model performance on small dataset $\mathbf{D_{10}}$

| Iteration | Type | Validation MSE Loss ($\times 10^{-3}$) | | | | Reduction |
|---|---|---|---|---|---|---|
| | | Model 1 | Model 2 | Model 3 | Average | |
| 0 | Trained Directly | 27.69 | 28.80 | 21.14 | 25.88 | - |
| 1 | PL | 20.05 | 20.30 | 20.95 | 20.43 | 21.04% |
| 2 | PL | 18.01 | 17.99 | 18.52 | 18.17 | 11.05% |
| 3-1 | PL | 17.24 | 17.84 | 17.40 | 17.49 | 3.76% |
| 3-2 | DAN-TR | 9.42 | 14.93 | 11.78 | 12.04 | 33.72% |
| 4 | PL | 7.07 | 6.98 | 6.76 | 6.94 | 42.42% |
| 5 | PL | 6.28 | 6.28 | 6.30 | 6.29 | 9.36% |
| 6 | DAN-TR | 4.78 | 4.89 | 4.83 | 4.83 | 5.08% |
| 7 | PL | 5.92 | 5.91 | 6.07 | 5.97 | 10.29% |
| 8 | PL | 5.36 | 5.36 | 5.34 | 5.35 | 6.61% |
| 9-1 | PL | 4.99 | 4.97 | 5.04 | 5.00 | 3.38% |
| 9-2 | DAN-TR | 4.74 | 4.56 | 4.72 | 4.68 | 6.45% |
| 10 | PL | 4.27 | 4.27 | 4.27 | 4.27 | 8.70% |
| 11-1 | PL | 4.17 | 4.20 | 4.20 | 4.19 | 1.84% |
| 11-2 | DAN-TR | 4.18 | 4.11 | 4.14 | 4.15 | 2.91% |
| 12 | PL | 3.96 | 3.99 | 3.98 | 3.98 | 4.06% |
| 13 | Final Training | 3.87 | 3.85 | 3.87 | 3.86 | 2.89% |

The following conclusions can be drawn from Fig. 5 and Table 2:

(1) The performance of models directly trained on the small-scale training set $\mathbf{D_{10}}$ is far from acceptable owing to data limitations. The average MSE is as high as $25.88\times10^{-3}$, nearly 30 times higher than that on $\mathbf{D_{320}}$ ($0.91\times10^{-3}$). Furthermore, there is a large variation in the performances of three obtained models, indicating that the training process is unstable.

(2) The network performance improves significantly after adopting the proposed self-transfer method, and the average MSE drops to $3.86\times10^{-3}$. The reduction ratio can reach 85.1%.

(3) The training process reveals that after repeatedly using the PL strategy (light green points in Fig. 5), the network performance improvement exponentially diminishes, confirming the theoretical analysis that the PL strategy cannot be used continuously.

(4) Following the execution of the PL strategy for one or two times, the DAN-

TR network is used to perform domain adaptation (red points in Fig. 5), which can not only improve the model performance in the current iteration, but also expand the potential for further improvement. Compared to the expected improvement attenuation of the PL strategy (as discussed in (3)), the relative loss reduction in the next iteration after DAN-TR increases significantly (for example, iteration 4 compared to iterations 2 and 3-1, and iteration 10 compared to iterations 8 and 9-1).

To further confirm the reliability of these conclusions, two small-scale datasets with 5 and 20 samples (denoted as $D_5$ and $D_{20}$, respectively) are randomly established based on $D_{320}$. Network training and validation are independently performed. Table 3 lists the validation results, and Fig. 6 shows the relative reduction of the MSE.

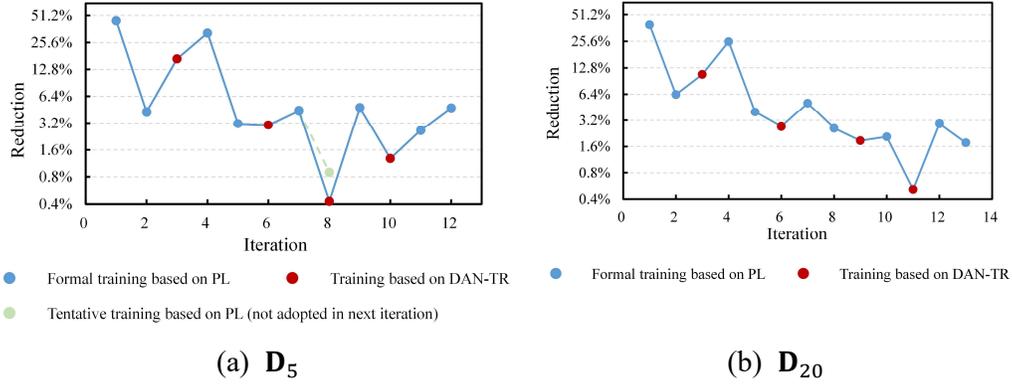

(a) $D_5$             (b) $D_{20}$

Fig. 6: Relative reduction of MSE loss in each iteration ($D_5$ & $D_{20}$)

Table 3: Model validation on $D_5$ and $D_{20}$

| Type | Average Validation MSE Loss ($\times 10^{-3}$) | |
|---|---|---|
| | $D_5$ | $D_{20}$ |
| Training Directly | 28.44 | 8.46 |
| Final Training Using the Proposed Method | 6.55 | 2.51 |
| Reduction | 77.0 % | 70.3 % |

The following conclusions can be drawn from Table 3 and Fig. 6:

(1) Model performance is unacceptable when the models are directly trained on small datasets. The average MSE on $D_5$ and $D_{20}$ are $28.44 \times 10^{-3}$ and $8.46 \times 10^{-3}$, respectively.

(2) Network performance significantly improves after using the self-transfer method proposed in this study. The average MSE on $D_5$ and $D_{20}$ drop to $6.55\times10^{-3}$ and $2.51\times10^{-3}$, with a reduction ratio of 77.0 % and 70.3 %, respectively. Models trained with only 5 samples outperform the models trained directly with 20 samples.

(3) Analysis of the training process supports the effectiveness of the proposed training framework. After using the PL strategy iteratively for one to two times, using DAN-TR can improve the model performance in the current iteration. Meanwhile, it significantly expands the potential of using the PL strategy in the following iterations.

### 4.4 Testing and visualization on small-scale datasets

The initial and final models on each dataset are chosen based on the validation results, and the performances of these models are assessed and compared based on the testing dataset. The results are summarized in Table 4. To intuitively show the differences in model performance with different MSEs, typical testing samples are visualized in Fig. 7. Clearly, the models trained using the proposed self-transfer method perform significantly better on the testing dataset. The MSE decreases by more than 70% on all small-scale datasets ($D_5$, $D_{10}$ and $D_{20}$), with the highest reduction ratio of 84.1%.

Table 4: Model performance on the testing dataset

| Type | Average Testing MSE Loss | | | |
|---|---|---|---|---|
| | $D_5$ | $D_{10}$ | $D_{20}$ | $D_{320}$ |
| Training Directly ($\times10^{-3}$) | 25.70 | 29.67 | 9.34 | |
| Final Training Using the Proposed Method ($\times10^{-3}$) | 5.69 | 4.71 | 2.78 | 0.92 |
| Reduction | 77.9 % | 84.1 % | 70.2 % | - |

The performance comparison of the final model trained on $D_{20}$ with that of the model trained directly on $D_{320}$ is visualized in Fig. 8. Despite that the amount of data differs by 16 times, the model performance is relatively similar, thus the requirements on datasets are significantly relaxed.

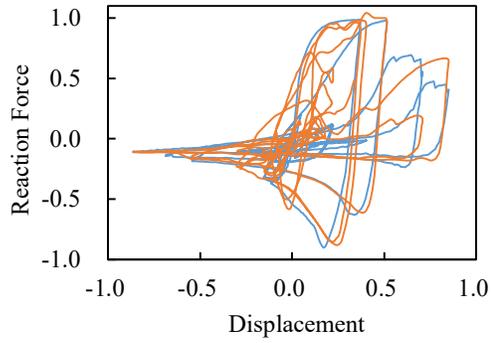

(a) **D₅**-Training directly
(MSE: 6.55×10⁻²)

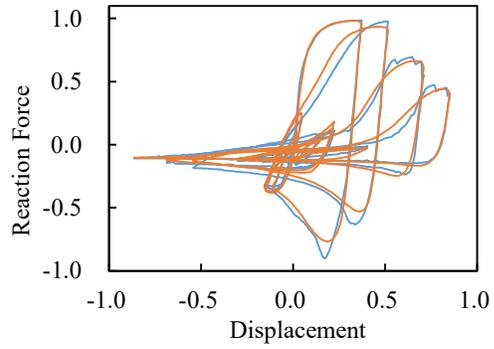

(b) **D₅**-Final training using the proposed
method (MSE: 1.12×10⁻³)

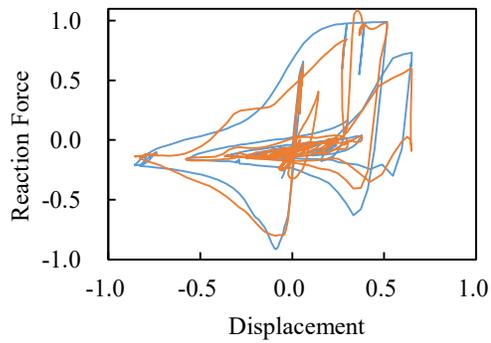

(c) **D₁₀**-Training directly
(MSE: 7.43×10⁻³)

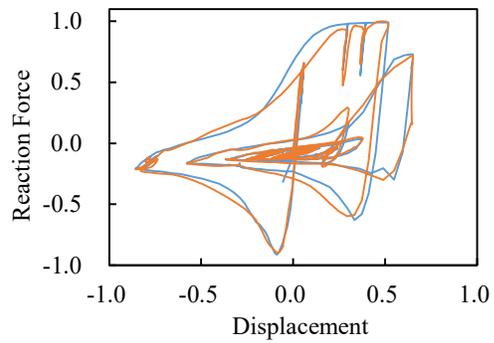

(d) **D₁₀**-Final training using the
proposed method (MSE: 1.31×10⁻³)

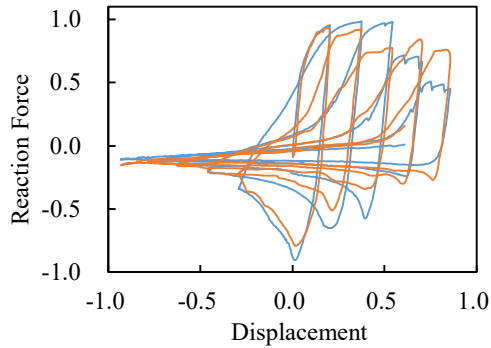

(e) **D₂₀**-Training directly
(MSE: 9.83×10⁻³)

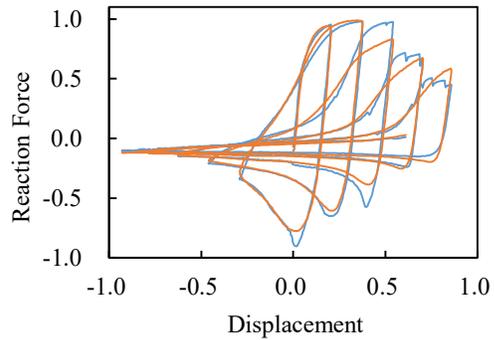

(f) **D₂₀**-Final training using the
proposed method (MSE: 2.61×10⁻³)

Fig. 7: Typical predictions of models trained on small datasets
(the blue lines represent the ground truth, and the orange lines represent the
prediction results)

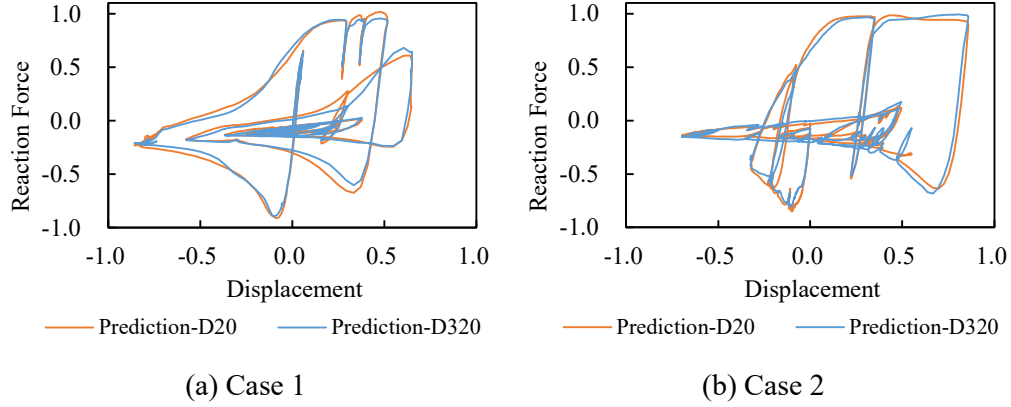

<table>
<tr><td>(a) Case 1</td><td>(b) Case 2</td></tr>
</table>

Fig. 8: Comparison of model performance ($\mathbf{D}_{20}$ and $\mathbf{D}_{320}$)

### 4.5 Method comparison

The proposed self-iterative training method is compared to the six types of widely recognized and adopted methods listed below. For all of the adopted methods, discussions on public hyperparameters and network architecture settings (such as learning rate, hidden dimensions, and number of hidden layers) are carried out, and pertinent discussions are conducted based on the private characteristics of each algorithm (more discussion details are explained below). All training/validation/testing processes are carried out on the same dataset (small dataset $\mathbf{D}_{10}$) to ensure comparability.

(1) Traditional data augmentation method

In this part, two randomly selected labeled sequences are sliced, spliced, or averaged with random weights. Augmented datasets with different sizes (100, 500, 1,000, and 5,000 samples) are generated. Then, LSTM networks with different hyperparameters are directly trained on those augmented datasets.

(2) Cross-stitch network

A well-known network architecture for multi-task learning is the cross-stitch network (Misra et al., 2016). Based on several "cross layers", the learning process of multiple tasks is coordinated with each other. In this study, the auxiliary task outputs are chosen as pseudo-labeled outputs with varying MSE loss (ranges from $6.92{\times}10^{-6}$ to $4.19{\times}10^{-2}$), and the inputs are the same (noiseless inputs). The number of auxiliary tasks discussed herein ranges from 1 to 4.

(3) Dynamically masked pre-train and fine-tune

This algorithm is proposed by Liu et al. (2019) and has been used in a number of other studies. During the pre-train stage, a certain percentage of the steps in the target sequences are masked (usually 15%), and the networks predict these masked values based on unmasked inputs. Following pre-training, the best models will be adopted and fine-tuned for the target task. To improve the model performance, the definition of pre-train tasks, masked sequence type (displacement/reaction force), and percentage of masked steps in the sequence are analyzed in this study.

(4) Semi-supervised learning method based on consistency loss

Consistency loss is widely adopted in semi-supervised learning (e.g., Laine & Aila, 2017; Tarvainen & Valpola, 2017; Xie et al., 2020). The main idea is to add a consistency loss to make the network more robust to noised input. To achieve better results, the weight of consistency loss (compared to regression loss) and the level of input noise are examined in this study. Furthermore, as a comparison method, the mean-teacher strategy (a widely accepted consistency-based method) (Tarvainen & Valpola, 2017) is combined with the PL strategy.

(5) Mixtext

Mixtext is a linguistically-informed interpolation method of hidden space for semi-supervised text classification (Chen et al., 2020b). It proposes a T-mix method to mix hidden representations (rather than input sequences) for data augmentation, which differs from traditional mixing methods (e.g., Mixup (Zhang et al., 2017)). Consistency loss is also adopted in the training process. In this study, LSTM is selected as the basic network of Mixtext, and the same unlabeled dataset established in Section 4.1 is used to provide unlabeled samples. To achieve better results, the type of mixed sequences (labeled/unlabeled/both), level of input noise, weight of consistency loss, and total number of unlabeled samples are all analyzed.

(6) Fine-tune based on well-known pre-trained models

This method is widely used in the field of time series regression, particularly in natural language processing (e.g., Devlin et al., 2019; Dai et al., 2019; Brown et al., 2020; Sun et al., 2021). In this study, several different pre-trained models (BERT-Tiny, BERT-Base-Uncased, BERT-Base-Chinese, and BERT-Large) are used as the foundation for fine-tuning to achieve better results. These models differ in terms of the pre-training source domain and the level of model complexity.

As shown in Table 5, the best results in each method category are chosen for comparison. The proposed iterative self-transfer learning method outperforms all other methods used for comparison.

Table 5 Comparison of different methods

| Method | Validation MSE ($\times 10^{-3}$) | Testing MSE ($\times 10^{-3}$) |
|---|---|---|
| Traditional Data Augmentation | 13.67 | 13.99 |
| Cross-stitch Network | 15.65 | 16.52 |
| Dynamically Masked Pre-train and Fine-tune | 9.57 | 10.54 |
| Semi-supervised Learning Method Based on Consistency Loss | 11.60 | 13.47 |
| PL Strategy & Mean Teacher | 17.17 | 17.99 |
| Mixtext | 13.11 | 14.42 |
| Fine-tune Based on Well-known Pre-trained Models | 124.08 | 125.45 |
| Proposed Iterative Self-transfer Learning Method | **3.85** | **4.69** |

## 4.6 Universality verification based on GA network

As discussed in Section 3.2, the simple LSTM could be used for dataset establishment (steps 1-4), and complex networks could be directly adopted in step 5. In this section, the complex GA network (Wang et al., 2020) is directly trained based on the final dataset generated through LSTM (including the labeled and pseudo-labeled samples). For comparison, GA network is also directly trained on small dataset ($\mathbf{D_5}$, $\mathbf{D_{10}}$ and $\mathbf{D_{20}}$). The validation and testing results are shown in Table 6.

According to Table 6, the dataset constructed based on the simple LSTM network and the proposed self-transfer method can also be used as the training basis for the complex GA network, and can bring significant performance improvements. Compared with training directly on small datasets, the MSE of the validation set decreases by 37.8% to 61.2%, and that of the testing set decreases by 16.5% to 56.7%. Therefore, datasets obtained by self-transfer training are versatile in different network architectures.

Table 6: Validation and testing results based on GA network

| Dataset | Type | Validation | | Testing | |
|---------|------|------------|-----------|---------|-----------|
| | | MSE | Reduction | MSE | Reduction |
| $\mathbf{D_5}$ | Training Directly | $1.91 \times 10^{-4}$ | 37.8 % | $1.78 \times 10^{-4}$ | 16.5 % |
| | Final Training Using the Proposed Method | $1.19 \times 10^{-4}$ | | $1.49 \times 10^{-4}$ | |
| $\mathbf{D_{10}}$ | Training Directly | $2.05 \times 10^{-4}$ | 61.2 % | $2.35 \times 10^{-4}$ | 56.7 % |
| | Final Training Using the Proposed Method | $7.95 \times 10^{-5}$ | | $1.02 \times 10^{-4}$ | |
| $\mathbf{D_{20}}$ | Training Directly | $1.16 \times 10^{-4}$ | 39.9 % | $1.43 \times 10^{-4}$ | 38.2 % |
| | Final Training Using the Proposed Method | $6.98 \times 10^{-5}$ | | $8.83 \times 10^{-5}$ | |
| $\mathbf{D_{320}}$ | Training Directly | $5.15 \times 10^{-5}$ | - | $5.35 \times 10^{-5}$ | - |

## 5. Conclusions

This study proposes a general method called "iterative self-transfer learning" for response time-history prediction. The following conclusions are drawn:

(1) The iterative self-transfer learning method proposed in this study can be used on small-scale target domain datasets, without the need to introduce large-scale source domain dataset or well behaved pre-trained models, add additional artificial labels, or perform complex physical/mathematical analyses. Datasets obtained by self-transfer training (with labeled and pseudo-labeled samples) are versatile in different network architectures. Therefore, the proposed method is valuable for many tasks with limited data.

(2) The case study demonstrates that using the proposed method on small-scale datasets (such as $\mathbf{D_5}$, $\mathbf{D_{10}}$, and $\mathbf{D_{20}}$) can significantly improve model performance. The testing MSE loss can be decreased by 70% to 84%. The prediction results of the final model trained on $\mathbf{D_{20}}$ are close to the performance of the models directly trained on $\mathbf{D_{320}}$, and are in good agreement with the real behaviors. Thus, the cost of data acquisition is drastically reduced.

(3) The DAN-TR network developed in this study can enhance domain adaptation and feature generalization. Training with the DAN-TR network can not only improve the model performance, but also expand the potential for performance improvement in subsequent PL strategy applications. Using a pseudo-labeled dataset as the source domain for transfer learning is a viable option.

(4) When using the PL strategy to improve the model performance, one to two iterations are sufficient. More iterations of the continuous use of the PL strategy will not result in a significant loss reduction.

Furthermore, even if the proposed method is used, this study shows that a larger dataset has advantages in terms of final performance, emphasizing the importance of data accumulation.

## Acknowledgement

This work is supported by the National Natural Science Foundation of China (No. 52238011), the National Key R&D Program (No. 2019YFC1509305) and the Tencent Foundation through the XPLORER PRIZE. The authors are grateful for Mingxuan Jing in Tsinghua University for giving constructive suggestions.